\documentclass[sigconf]{acmart}
\usepackage{amsmath,graphicx,url,multirow}
\usepackage{xcolor}

\AtBeginDocument{%
  \providecommand\BibTeX{{%
    \normalfont B\kern-0.5em{\scshape i\kern-0.25em b}\kern-0.8em\TeX}}}

\copyrightyear{2021}
\acmYear{2021}
\setcopyright{rightsretained}
\acmConference[MMSports '21]{Proceedings of the 4th International Workshop on Multimedia Content Analysis in Sports}{October 20, 2021}{Virtual Event, China}
\acmBooktitle{Proceedings of the 4th International Workshop on Multimedia Content Analysis in Sports (MMSports '21), October 20, 2021, Virtual Event, China}\acmDOI{10.1145/3475722.3482793}
\acmISBN{978-1-4503-8670-8/21/10}

\begin{document}
\fancyhead{}

\title[Three-Stream 3D/1D CNN for Fine-Grained Action Classification and Segmentation in Table Tennis.]{Three-Stream 3D/1D CNN for Fine-Grained Action Classification and Segmentation in Table Tennis.}





\author{Pierre-Etienne Martin}
\orcid{0000-0002-9593-4580}
\affiliation{%
  \institution{Max Planck Institute for Evolutionary Anthropology}
  \streetaddress{Deutscher Platz 6 }
  \city{D-04103 Leipzig}
  \country{Germany}
}
\email{pierre_etienne_martin@eva.mpg.de}

\author{Jenny Benois-Pineau}
\affiliation{%
  \institution{Univ. Bordeaux, CNRS,  Bordeaux INP, LaBRI, UMR 5800}
  \city{F-33400, Talence}
  \country{France}
  }
\email{jenny.benois-pineau@u-bordeaux.fr}

\author{Renaud P\'{e}teri}
\affiliation{%
  \institution{MIA, La Rochelle University}
  \city{La Rochelle}
  \country{France}}
\email{renaud.peteri@univ-lr.fr}

\author{Julien Morlier}
\affiliation{%
  \institution{IMS, University of Bordeaux}
  \city{Talence}
  \country{France}
}








\begin{abstract}
This paper proposes a fusion method of modalities extracted from video through a three-stream network with spatio-temporal and temporal convolutions for fine-grained action classification in sport. It is applied to \texttt{TTStroke-21} dataset which consists of untrimmed videos of table tennis games. The goal is to detect and classify table tennis strokes in the videos, the first step of a bigger scheme aiming at giving feedback to the players for improving their performance.
The three modalities are raw RGB data, the computed optical flow and the estimated pose of the player. The network consists of three branches with attention blocks. Features are fused at the latest stage of the network using bilinear layers. Compared to previous approaches, the use of three modalities allows faster convergence and better performances on both tasks: classification of strokes with known temporal boundaries and joint segmentation and classification. The pose is also further investigated in order to offer richer feedback to the athletes.
\end{abstract}



\begin{teaserfigure}
    \centering
    \includegraphics[width=.24\linewidth]{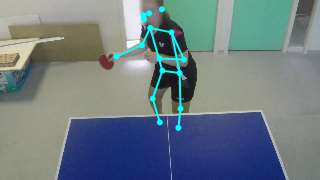}
    \includegraphics[width=.24\linewidth]{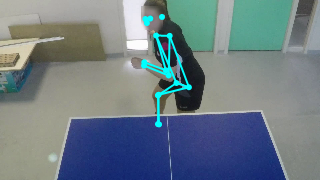}
    \includegraphics[width=.24\linewidth]{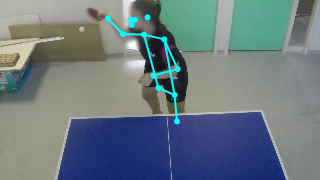}
    \includegraphics[width=.24\linewidth]{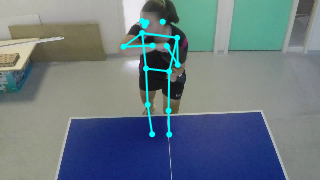}\\
    \includegraphics[width=.24\linewidth]{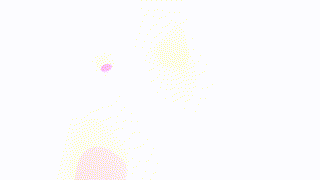}
    \includegraphics[width=.24\linewidth]{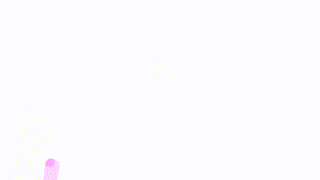}
    \includegraphics[width=.24\linewidth]{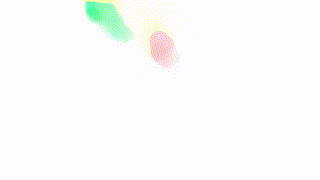}
    \includegraphics[width=.24\linewidth]{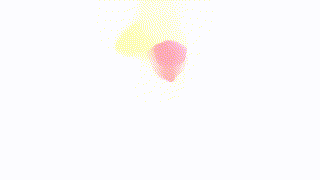}\\
  \caption{Frames of an ``Offensive Forehand Hit'' stroke from \texttt{TTStroke-21} with its estimated pose and optical flow.}
  \label{fig:teaser}
\end{teaserfigure}

\maketitle

\section{Introduction and related works}
\label{sec:intro}

\par
Fine-grained action classification is being more and more investigated in recent years due to its various potential applications such daily living care~\cite{Dataset:ADLEgoDataset_long, Dataset:ToyotaSmarthome:2019_long}, video security and surveillance~\cite{NN:MultiStream:2016_long} or in sport activities~\cite{PeMTAP:2020_long,Dataset:Gym:2020_long,Dataset:Diving48:2018}. The difference with coarse-grained action classification~\cite{Dataset:Kinetics700:2020,Dataset:AVA_Kinetics:2020_long,Dataset:UCF101:2012} lays in the high intra-class similarity of the actions. Movements performed are often similar since they focus on one particular activity. Moreover, since videos are recorded in the same context, the background scene and manipulated objects are similar in all videos. Consequently, all possible information should be extracted from the performed movement itself in order to discriminate actions. The target application of our research is fine-grained action recognition in sports with the aim of improving athletes performance.
\par
Collecting individual data from athletes by body-worn sensors (connected watches, smart clothes, exoskeleton) might be a valuable source of information for classification of similar actions. However, the analysis of gestures is often confined to laboratory studies ~\cite{Sports:TennisStrokeClassification:2019_long,TT:RecognitionWithSensor:2019_long,TT:RecognitionWithSensorHybrid:2020_long}. Sound has also proven to be efficient for event detection~\cite{DBLP:conf/mm/BaughmanMRGHW19} but may not be used for more complex tasks. In~\cite{TT:TTNet:2020}, the authors propose an advanced real-time solution for scene segmentation, ball trajectory estimation and event detection but are not considering stroke classification.
\par
The use of pose, expressed as coordinates of skeleton joints, has also become popular for action recognition. In~\cite{NN:Wu:2016_long}, the authors apply 3D CNNs on gesture recognition with RGB-D data. They use only four frames mixing 3D and 2D convolutions and max pooling. Joint information is fused using a Deep Belief Network~\cite{Classification:DBN2:2016_long}. Similarly, PoTion~\cite{Potion:2018_long} uses movement of the human joints as features to improve the classification score of the I3D models~\cite{NN:I3DCarreira:2017_long}. Pose has also been used in sport: \cite{Sport:RecognitionSoccerPose:2019_long}~performs classification of four classes of football (soccer) actions based on pose estimation. The authors of~\cite{DBLP:conf/mm/ShimizuHSYL19} investigate shot direction in Tennis using the pose of the player. In~\cite{Pose:ActionRecognitionAndPoseEstimation:2018_long}, the authors propose a multi-task method for 2D/3D pose estimation and action recognition. Similarly, \cite{Pose:LCRNet++:2020}, based on LRC-Net~\cite{Pose:LCRNet:2017}, builds a pseudo ground truth for 3D poses from images using 2D pose search in a projected 3D pose dataset in order to offer 3D human pose from images. In \cite{Pose:RepresentationForClassification}, pose representations from the pose estimators are feed to a 3D CNN in order to obtain spatio-temporal representation used for action classification. Furthermore, a 3D attention mechanism has been investigated on joint skeleton using LSTM~\cite{Attention::skeleton_long}. The pose can also be used for spatial segmentation as in~\cite{ActionRecognition:BoxOfPose:2019_long}.
\par
However, there are limitations of using skeleton-based approaches for action recognition as pointed out in~\cite{Pose:Vulnerability:2020_long}. The authors manage to induce large errors with attacks on the pose based models through low variation of the inputted pose. The use of several modalities is therefore needed to be less dependent on the pose estimation alone. Pose can also be used for further analysis: in~\cite{TT:TrajectoryUsingPose:2020_long} the position of a table tennis ball is predicted according to the player's pose. In~\cite{Sport:EvaluationMotion:2017_long}, qualitative measures of tennis and karate gestures are computed for comparing the pose of expert and novice participants.

\par
Recording of ``markerless'' and ``sensorless'' video of performing athletes has an advantage. It does not bias human performance in the target task. In this case the classification of actions has to be done using video only. Hence, as much as possible information must be extracted from the video stream in order to conduct movement analysis. The first modality is the raw information of pixel colour values. Motion is an important modality, extracted by optical flow, as investigated in~\cite{PeICIP:2019_long}. It was proved to be efficient in terms of classification performance. Improvements can be achieved by making use of other information from the video recordings, such as the sound. Another possibility is to extract an information which is the interpretation result of raw data. Thus, we consider a human pose expressed via joints spatial coordinates which can be computed from the same videos. The purpose is to make cameras ``smart'' to analyse sport practices~\cite{LenhartL:2018_long}.
\par
In this work, we investigate the use of pose information for classification inspired by the work carried out in~\cite{PeMTAP:2020_long,PeICPR:2020_long}.
In the original twin model that takes as input RGB stream and its estimated optical flow, a third branch with pose information is added to the network. The branches are fused at the latest stage of the network through several bilinear layers. Experiments are performed on the \texttt{TTStroke-21} dataset. We solve two distinctive tasks: classification only and joint classification and segmentation from videos. Our method achieves slightly better performance on the classification task but much better scores on the joint classification and segmentation task through the use of  pose and a fusion approach. We also present the opportunities that the pose offers for further movement analysis to enrich the feedback to the users.
\par
The paper is organised as follows: computed modalities and classification method are introduced in section~\ref{sec:Approach}. Results and other potential applications of the pose information are discussed in section~\ref{sec:results}. Conclusion and prospects are drawn in section~\ref{sec:conclusion}.

\section{Proposed approach}
\label{sec:Approach}
\begin{figure*}[tbp]
	\centering
	\includegraphics[width=\linewidth]{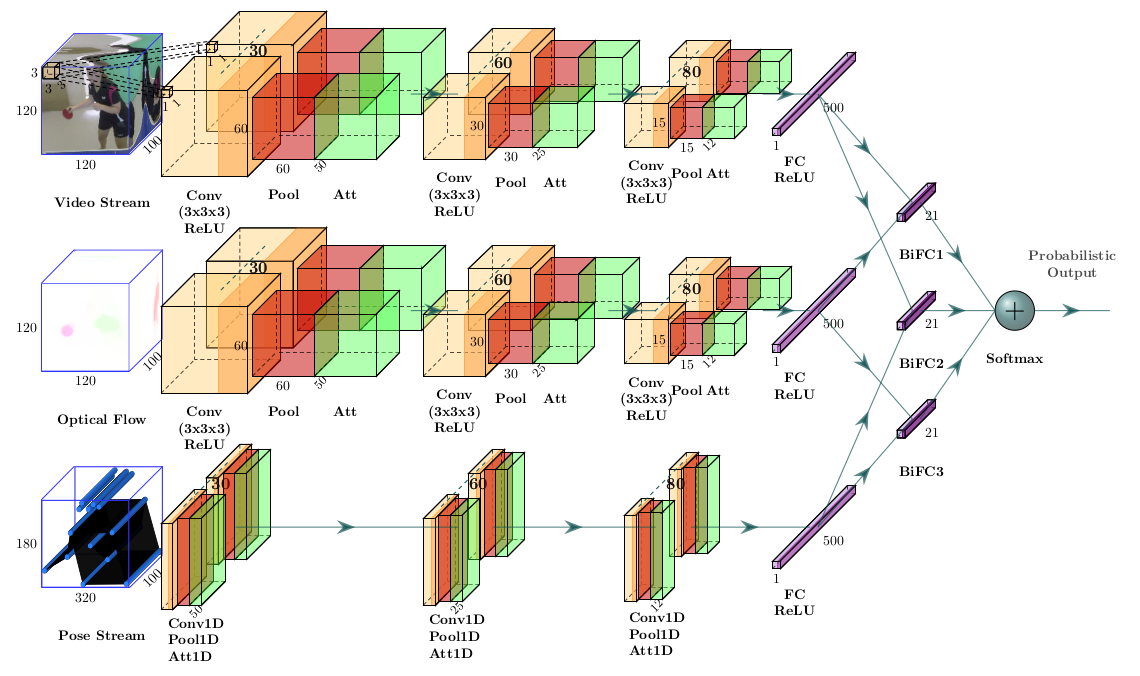}
	\caption{Three-Stream architecture processing RGB, optical flow and pose data in parallel with spatio-temporal convolutions.}
	\label{fig:architecture}
\end{figure*}

To deal with the low inter-class variability of the actions proper to fine-grained action classification, the most complete information from video must be extracted, i.e. both appearance (RGB) and motion (Optical Flow) modalities. Spatio-temporal convolutions in the network are performed on cuboids of RGB frames and on cuboids of optical flow (OF). Pose joints are also processed by temporal convolutions. All three modalities are processed simultaneously through a three-stream architecture as presented in Figure~\ref{fig:architecture}.

\subsection{Optical Flow Estimation}
\label{section:flows}
As presented in~\cite{PeICIP:2019_long}, the OF and its normalization can strongly impact the classification results. The same motion estimator reaching best classification performances is used thereafter. The method is based on iterative re-weighted least square solver~\cite{OF:BP:2009_long}. Each OF frame $V=(v_x,v_y)$ is encoded with its horizontal $v_x$ and vertical $v_y$ motion components being computed from two consecutive RGB frames. In order to only keep foreground motions, estimated OF is smoothed with Gaussian filter with kernel size $3 \times 3$ and multiplied using Hadamard product by the computed foreground mask $M_{FG}$: $V_{FG}=V\odot (M_{FG},M_{FG})$~\cite{Zivkovic:2006_long}.

\subsection{Region Of Interest Estimation}
\label{subsection:Spatial_segmentation}

The region of interest (ROI) center ${\bf X_{roi}} = (x_{roi},y_{roi})$ is estimated from the maximum of the OF norm and the center of gravity of non-null OF values as follows:
\begin{equation}
\label{Chap4:eq:Region-of-Interest extraction}
	\begin{array}{l}
		{\bf X_{max}} = (x_{max}, y_{max}) =\underset{x, y}{argmax}(||{\bf V}||_1)\\
        {\bf X_g} = (x_{g}, y_{g}) = \dfrac{1}{\underset{{\bf X} \in \Omega}{\sum{\delta({\bf  X})}}}
     			\underset{{\bf X} \in \Omega}{\sum{{\bf X}\delta({\bf X})}}\\
        \mbox{with} ~~~ \delta({\bf X}) = \left\{
                            \begin{array}{ll}
                                1 & \mbox{if } ||{\bf V({\bf X})}||_1 \neq 0 \\
                                0 & \mbox{otherwise}
                            \end{array}
                            \right. \\
		x_{roi} = \alpha~f_{\omega_x}(x_{max},~ W) + (1-\alpha)~ f_{\omega_x}(x_{g},~ W)\\
		y_{roi} = \alpha~f_{\omega_y}(y_{max},~ H) + (1-\alpha)~ f_{\omega_y}(x_{g},~ H)
	\end{array}
\end{equation}

with parameter $\alpha$ set empirically to $0.6$,  ${\bf \Omega} = (\omega_x, \omega_y ) = (320, 180)$ the size of video frames. Function $f_{\omega}(u,S) = max(min(u,\omega-\frac{S}{2}), ~\frac{S}{2})$ allows to define ROI without the image border. The size of cuboids are  $(W \times H \times T) = (120 \times 120 \times 100)$ which corresponds to a duration of 0.83s. To avoid jittering within the RGB and OF, a Gaussian filter of kernel size $k_{size}=41$ ($\frac{1}{3}$~second) and scale parameter $\sigma_{blur} = 0.3*((k_{size}-1)*0.5-1)+0.8$ is applied along the temporal dimension to average the ROI center position. These parameter values were chosen experimentally and are suitable for the 120 fps video frame rate. 

\subsection{Pose Estimation}
The pose is computed from single RGB frames using the PoseNet model~\cite{Pose:PersonLab:2018_long}. Its implementation is available online\footnote{\url{https://github.com/rwightman/posenet-python}}.
It supplies poses and human joints positions and their confidence score. In addition, the pose position (mean of the joint coordinates) and its attributed score are used, leading to a descriptor vector $J$ with elements such as:
\begin{equation}
    \label{eq:joint}
    J(i)=(x_i,y_i,s_i)^T
\end{equation}
with $i$ the $i^{th}$ joint or the pose, $x_i$  and $y_i$ its horizontal and vertical coordinate and $s_i$ its associated score.

\subsection{Data Normalization}
To map their values into interval $[0,1]$, RGB data are normalized by theoretical maximum, while joints position $x_i$ and $y_i$ are normalized with respect to the width and height of the video frames: $W_{img}$ and $H_{img}$. The OF is normalized using the mean $\mu$ and standard deviation $\sigma$ of the maximum absolute values distribution of each OF components over the whole dataset as described in equation~\ref{eq:OFnormalization}:

\begin{equation}
\label{eq:OFnormalization}
	\begin{array}{l}
	v' = \frac{v}{\mu + 3\times \sigma}\\
	v^N(i,j) = \left\{
	\begin{array}{ll}
	v'(i,j) & \mbox{if } |v'(i,j)| < 1 \\
	SIGN(v'(i,j)) & \mbox{otherwise.}
	\end{array}
	\right.
	\end{array}
\end{equation}

with $v$ and $v^N$ representing respectively one component of the OF $V_{FG}$ and its normalization. This normalization scales values into [-1,1] and increases the magnitude of most of vectors making the OF more relevant for classification.

\subsection{Model Architecture}
\label{subsection:architecture}

The architecture is inspired from the Twin Spatio-Temporal Convolutional Neural Network - T-STCNN with attention mechanisms presented in~\cite{PeICPR:2020_long} which takes as input the OF and RGB values through two branches using 3D convolutions and attention mechanism.
Compared to the latest, our network has three branches and takes as additional input joint coordinates. 
Furthermore the fusion step is adapted to fuse the three modalities.
\par
As depicted in Figure~\ref{fig:architecture}, the networks perform 3D (spatio-temporal) and 1D (temporal) convolutions. The two first branches are composed of three 3D convolutional layers with $30$, $60$, $80$ filters respectively which can be described by equation~\ref{eq:conv}:
 \begin{equation}
 \label{eq:conv}
    out(j) = bias(j) + \sum_{k=0}^{C_{in}-1} weight(j,k)\star X(k)
\end{equation} 
where $j$ is the $j^{th}$ output channel, $C_{in}$ the number of channels of the input $X$ and $\star$ is the valid 3D cross-correlation operator. 
Each branch takes cuboids of RGB values and OF of size $(W \times H \times T)$ with respectively $3$ and $2$ channels. The 3D convolutional layers use $3 \times 3 \times 3$ space-time filters with a dense stride and padding of $1$ in each direction. Their output is processed by max-pooling layers using kernels of size $2 \times 2 \times 2$. Each max-pooling layer feeds an attention block. The output of the successive convolutions is then flattened to feed a fully connected layer: $y = xA^T+b$ of length $500$.
\par
An extra branch processes the pose data of length $T$. It follows the same organization than the two other branches but uses 1D temporal convolutions over all the joints coordinates and scores (see eq.~\ref{eq:joint}) at the first convolution leading to $N_{joints}\times 3$ channels. This operation is similar to equation~\ref{eq:conv} using simple cross-correlation. A max-pooling operation is performed along the temporal dimension.
\par
The three branches are fused two by two using bilinear fully connected layers: $y=x_1^TAx_2+b$, of length $N_{classes}$, which represents the number of classes. The three resultant outputs are summed and processed by a Softmax function to output probabilistic scores used for classification.

\subsection{Data Augmentation}

Data augmentation is performed on-the-fly on the train set. Each stroke sample is fed to the model once per epoch. For temporal augmentation, $T$ successive data from the RGB, OF and Pose modalities, are extracted following a normal distribution around the center of the stroke video segment with a standard deviation of $\sigma= \frac{\Delta t-T}{N}$ with $N=6$. Spatial augmentation is performed with random rotation in range $\pm10^\circ$, random translation in range $\pm0.1$ in $x$ and $y$ directions, random homothety in range $1 \pm 0.1$ and flip in horizontal direction with $0.5$ of probability. The OF and Pose values are updated accordingly. Transformations are applied on the region of interest to avoid inputting regions outside the image borders.
During the test phase, no augmentation is performed and the $T$ extracted frames are temporally centered on the stroke segment.

\subsection{Training Phase}

All models are trained from scratch using stochastic gradient descent with Nesterov momentum and weight decay. Cross-entropy loss is used as objective function. A learning rate scheduler is used, which reduces and increases the learning rate when the observed metric (validation loss) reaches a plateau. Warm restart technique~\cite{Deep:LrDecrasedThenIncreased:2017} is used: weights and state of the model are saved when performing the best (lowest validation loss) and re-loaded when the learning rate is updated. This allows to re-start the optimization process from the past state with a new learning rate in the gradient descent optimizer.
\par
The following parameters were found optimal after successive experimental trials using grid search. Grid search was used for the following parameters: $start learning rate$, $patience$ and the number of epochs considered for comparing the training loss averages.
\par
Training process starts with a learning rate of $0.01$.	A number of epochs: $patience$, set to $50$, is considered before updating the learning rate, unless the performance drastically dropped (in our case: $0.7$ of the best validation accuracy obtained).
\par
The metric of interest is the training loss: if its average on the last $25$ epochs is greater than its average on the $35$ epochs before, the process is re-started from the past state and the learning rate divided by ten until reaching $10^{-5}$. After this step, the learning rate is set back to $0.01$ and the process continues.
This technique differs from decreasing only by step~\citep{Deep:LrDecreasedByStep:2016} since the learning rate might re-increase if no amelioration is observed.

\section{Experiments and Results}
\label{sec:results}

We compare results with the original T-STCNN with attention mechanism from~\cite{PeICPR:2020_long} and the Two-Stream I3D model~\cite{NN:I3DCarreira:2017_long}, all trained and tested from scratch on \texttt{TTStroke-21} (Fig.~\ref{fig:dataset}), and fed with cuboids of same size. As an ablation study, the three-stream network is trained with and without attention mechanism on the third branch, using in both cases a momentum of $0.5$, a weight decay of $0.05$ and a batch size of $5$ over $1500$ epochs. The learning rate varies between $0.01$ and $0.00001$. The two tasks are considered:  i) pure classification and ii) joint classification and segmentation. We also widen the field of application by considering the pose estimation for movement analysis.

\subsection{TTStroke-21 Dataset}
	
\begin{figure}[tb]
	\centering
	\begin{tabular}{cc}
	    \centering
        \includegraphics[height=.33\linewidth]{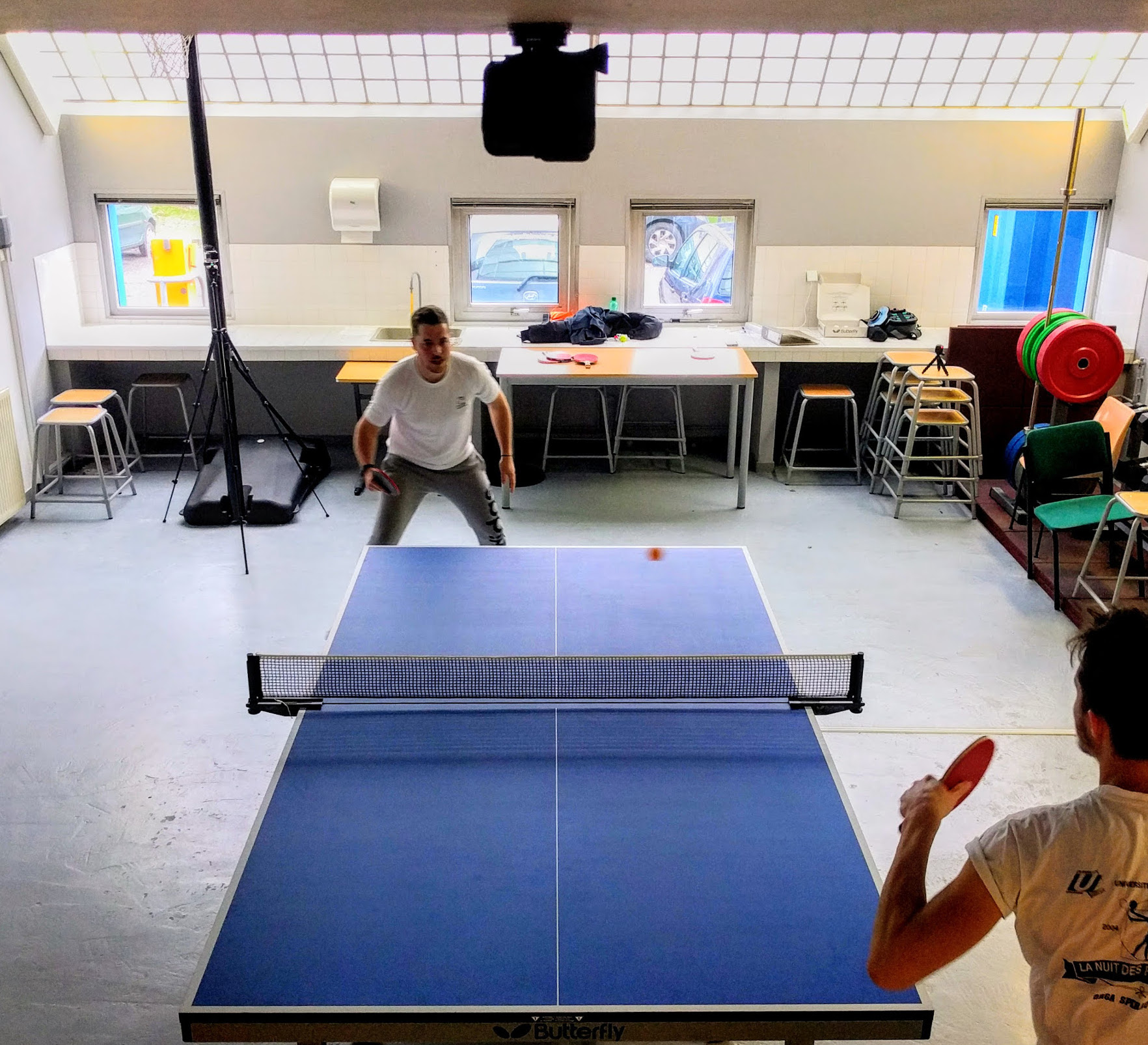}&
        \includegraphics[height=.33\linewidth]{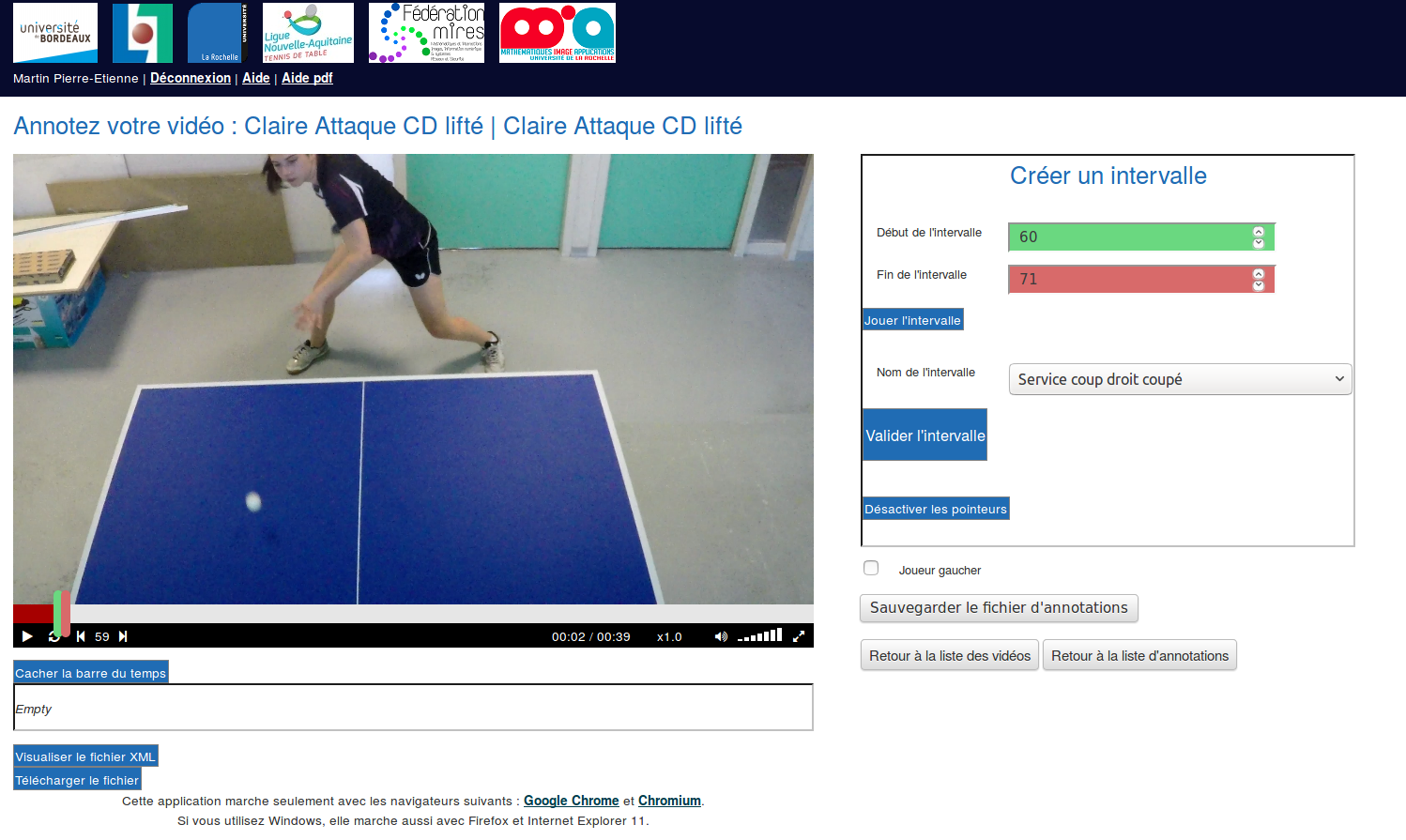}\\
        (a) Acquisition Setup.&
        (b) Annotation Platform.\\[3pt]
    \end{tabular}
    \includegraphics[width=.32\linewidth]{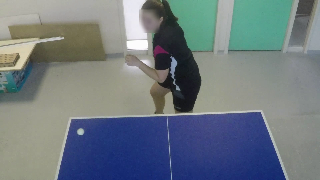}
    \includegraphics[width=.32\linewidth]{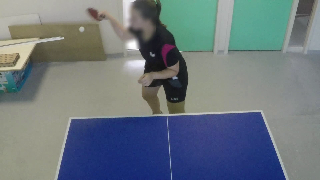}
    \includegraphics[width=.32\linewidth]{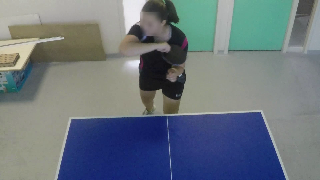}\\
    (c) Samples of a Stroke
    \caption{\texttt{TTStroke-21} Dataset.}
	\label{fig:dataset}
\end{figure}

\texttt{TTStroke-21}, depicted in Figure~\ref{fig:dataset}, is composed of table tennis videos, recorded indoors at different frame rates. The players are filmed in game or training situations, performing in natural conditions without marker. The videos have been annotated by table tennis players and experts from the Faculty of Sports (STAPS) of the University of Bordeaux, France. The number of classes considered is $N_{classes}=21$: \begin{itemize}
    \item \textbf{8 services:}
\textit{Serve Forehand Backspin,
Serve Forehand Loop,
Serve Forehand Sidespin,
Serve Forehand Topspin,
Serve Backhand Backspin,
Serve Backhand Loop,
Serve Backhand Sidespin,
Serve Backhand Topspin}; 
    \item \textbf{6 offensive strokes:}
\textit{Offensive Forehand Hit,
Offensive Forehand Loop,
Offensive Forehand Flip,
Offensive Backhand Hit,
Offensive Backhand Loop,
Offensive Backhand Flip;}
    \item \textbf{6 defensive strokes:}
\textit{Defensive Forehand Push,
Defensive Forehand Block,
Defensive Forehand Backspin,
Defensive Backhand Push,
Defensive Backhand Block,
Defensive Backhand Backspin};
\end{itemize}
and an extra \textit{negative} class.
\par
In the following experiments, $129$ of videos recorded at $120$ fps are used. They represent a total of $1048$ actions/strokes. From these temporally segmented table tennis strokes, $106$ negative additional samples are extracted from the rest of the videos. A larger number of negative samples could have been extracted, but this choice was made to have a lighter class imbalance, speed up the training process and be consistent with the previous experiments. The dataset is distributed in Train, Validation and Test sets with $0.7$, $0.2$ and $0.1$ proportions as in~\cite{PeMTAP:2020_long}. Extracted frames of size $(1920 \times 1080)$, are resized to $(W_{img}\times~H_{img})=(320\times 180)$ before computing modalities.
\par
Note that all computed human joints are not considered in pose data. Some of them might not be visible in the videos, e.g. knees and the ankles, which are often hidden by the table. We consider $13$ human joints: nose, eyes, ears, shoulders, elbows, wrists and hips. This leads to a descriptor $J$ of length $N_{J}=14$. Furthermore, other players may appear in the scene, which leads to the detection of several poses in the same frame. In this case, only the closest pose from the previously computed ROI center is considered. If no pose is detected (which is the case for 25\% of the frames), the descriptor vector is filled with ROI center coordinates and a score of $0$. Miss-detection of the pose happens in situations when the player is out of the camera field of view. This happens when the ball leaves the table and the player has to catch it, or at the beginning and at the end of the game.

\subsection{Pure Classification Task}

Results are compared with different models that have been tested following an ablation method~\cite{PeICPR:2020_long}. As it can be observed from table~\ref{table:acc}, the I3D model performances are worse compared to the others. The network is too deep for such a limited real-life dataset and the challenging fine-grained task. Furthermore, the classification performances of the Twin model and the Three-Stream model are similar. However, room for improvement still remains on the Three-Stream Network using attention mechanism since the gap between validation and train accuracy is lower. Convergence is achieved at epoch $1 176$ for the latest model after only $736$ effective epochs (counting only epochs following and saving of the state). The other models reach convergence only after epoch $1 400$.
\par
Preliminary results also have shown that the Pose alone could not achieve convergence and was able to classify with $22\%$ of accuracy only on the test set for the pure classification task. The importance of the combination of different information sources is thus obvious.

\begin{table}
	\centering
	\caption{Classification Performance in terms of Accuracy \%.}
	\label{table:acc}
	\begin{tabular}{cccc}
	\hline
    	Models & Train & Validation & Test \\
     	\hline
        RGB - I3D~\cite{NN:I3DCarreira:2017_long}         &$98$   &$72.6$ & $69.8$\\
        RGB-STCNN~\cite{PeICIP:2019_long} 		                &$98.6$	&$87$	&$76.7$\\
     	RGB-STCNN$^\dagger$	&$96.9$	&$88.3$	&$85.6$\\[3pt] 
    	
     	Flow - I3D~\cite{NN:I3DCarreira:2017_long}        &$98.9$ &$73.5$ & $73.3$\\
     	Flow-STCNN~\cite{PeICIP:2019_long}  	            &$88.5$	&$73.5$	&$74.1$\\
     	Flow-STCNN$^\dagger$ 	            &$96.4$	&$83.5$	&$79.7$\\[3pt]
    	
    	RGB + Flow - I3D~\cite{NN:I3DCarreira:2017_long}  &$99.2$     &$76.2$     &$75.9$\\
     	Twin-STCNN                                  &$99$	&$86.1$	&$81.9$\\
    	Twin-STCNN$^\dagger$                      &$97.3$	&$87.8$	&$87.3$ \\[3pt]
    	Three-Stream Net.*              &$97$   &$90$   &$87.3$ \\
    	\textbf{Three-Stream Net.$^\dagger$}     &$95.8$ &$86.5$ &$87.3$ \\
    	\hline
    \end{tabular}
        \\ $\dagger$ using attention mechanism on all branches
        \\ * using attention mechanism only on the OF and RGB branches 
\end{table}

\subsection{Joint Classification and Segmentation Task}

Similarly to \cite{PeMTAP:2020_long}, joint classification and segmentation of video segments is performed using an overlapping sliding window. Different decisions are investigated to flatten the obtained probabilities along the temporal dimension using a window size of $150$ for ``Vote'' and ``Avg'' rules, and size $201$ for ``Gauss'' rule. Once more, these window sizes were fixed after a preliminary grid search. The decision rules were respectively: i) majority vote, ii) average decision rule, and iii) weighted decision fusion using a Gaussian kernel. Performances are reported with all the labels, and also when the negative class is not considered. This second evaluation is motivated by the fact that most parts of a video are constituted of negative labels. Indeed, all portions between stokes are considered as negative: i.e. when the player is getting ready, when the match or training session end, and when the player is resting.

\begin{table}
    \centering
    \caption{Performance of Stroke Detection and Classification.}
    \label{table:acc_videos}
    \begin{tabular}{ccccc}
        \cline{2-5}
        & \multicolumn{4}{c}{Accuracies} \\
        \hline
        Models & Gross & Vote & Avg & Gauss \\
        \hline
        T-STCNN$^\dagger$       &$31$       &$46.8$	    &$47.7$     &$47.3$ \\
        \textbf{Three-Stream Net.*}            &$43.6$   &$63.1$   &$\bf63.9$ &$62.9$\\
    	Three-Stream Net.$^\dagger$  &$37.3$ &$57.9$ &$59$ &$57.8$ \\
        \hline
        \multicolumn{5}{c}{\textit{without taking into account the negative labels}}    \\
        \hline
        T-STCNN$^\dagger$   &$45.2$     &$63.8$	    &$65.6$     &$67.9$     \\
        \textbf{Three-Stream Net.*}            &$69.6$   &$83.7$   &$84.3$ &$\bf85.6$\\
    	Three-Stream Net.$^\dagger$ &$66.6$ &$82.1$ &$82.8$ &$84.1$ \\
        \hline
    \end{tabular}
        \\ $\dagger$ using attention mechanism on all branches
        \\ * using attention mechanism only on the OF and RGB branches 
\end{table}

Superiority of the Three-stream network is better observed for this joint classification and detection task, see table~\ref{table:acc_videos}. On the first part of the table, the Three-Stream Net is able to reach $63.9\%$ of accuracy against $47.7\%$ for the Twin model. Frame wisely, this represents a precision of $0.99$ and a recall of $0.42$ with regards to the negative class, leading to a F-score of $0.59$. This means the model is still more likely to classify as a stroke some frames belonging to a the negative class. However this score is also biased by the frame wised approach of the evaluation. This is why the second part of the table is of better interest: the add of the third branch allows to boost the performance up to $18$\% compared to the model without pose information. The attention mechanism performs slightly lower, which might be overcome with a longer training phase as observed earlier. Overall, better performances are noticed for all models when not considering the negative samples, which can be more challenging to classify. This may be due to all different and nonconventional gestures when a player attempts to catch a lost ball, leading to features similar to a stroke and classified as such.

\subsection{Perspectives for Movement Analysis}

\begin{figure}[t]
	\centering
	\begin{tabular}{cc}
		\includegraphics[width=0.46\linewidth]{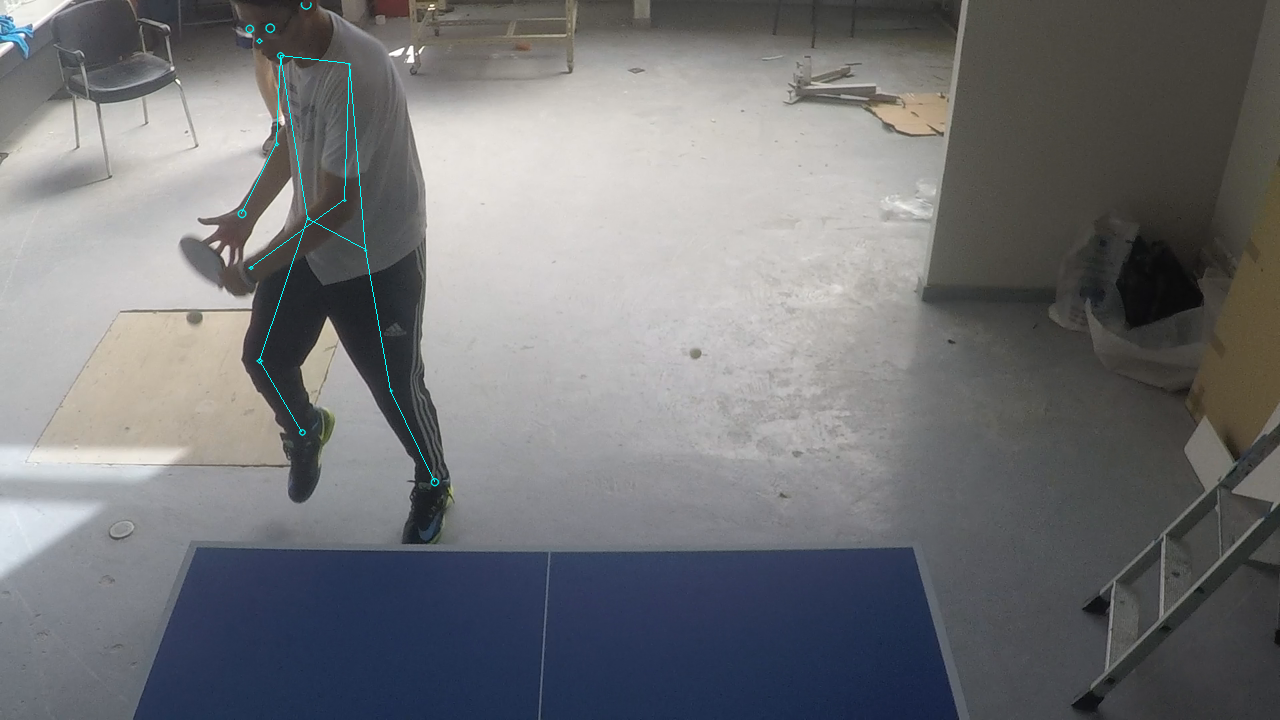}&
		\includegraphics[width=0.46\linewidth]{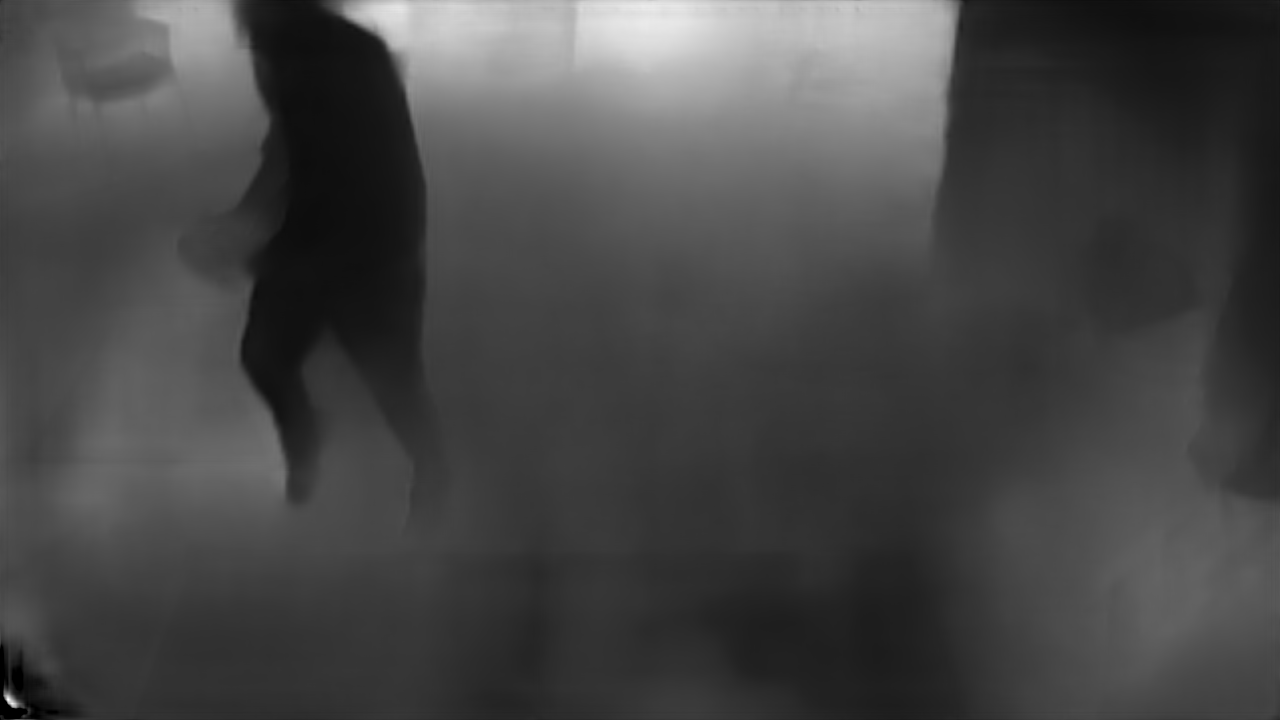}\\
		(a) Estimated pose \cite{Pose:PersonLab:2018_long} &
		(b) Estimated depth	\cite{Segmentation:SharpNetDepth:2019_long}\\[4pt]
	\end{tabular}
	\begin{tabular}{ccc}
		\includegraphics[height=0.42\linewidth]{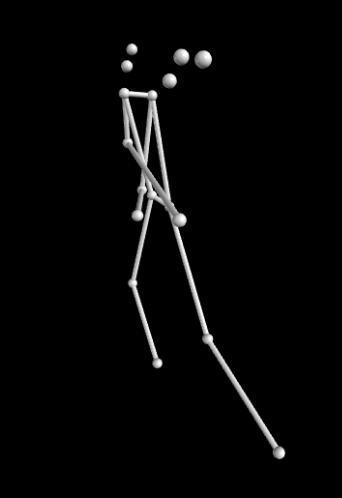}&
		\includegraphics[height=0.42\linewidth]{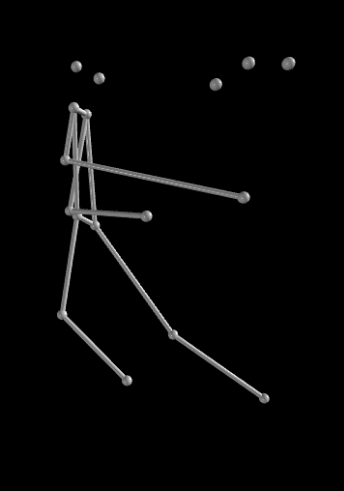}&
		\includegraphics[height=0.42\linewidth]{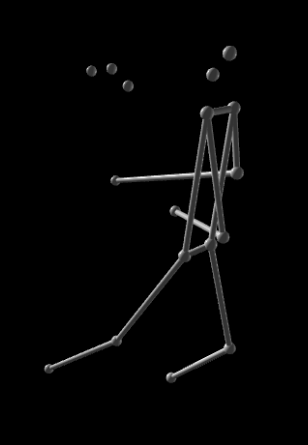}\\
		(c) Front &
		(d) Side &
		(e) Back sided\\
	\end{tabular}
	\caption{Combination of the pose and depth estimated from a single image (first row) to build a 3D skeleton (second row).}
	\label{fig:3dpose}
\end{figure}

Pose information may also be very interesting to assess the player's performance and the efficiency of his/her gesture. The organization of the joints skeleton, during a movement can be compared with a baseline to give an appreciation of the stroke performed~\cite{Sport:EvaluationMotion:2017_long}. Richer feedback could also be given by adding depth information, computed from a single image, in order to create a 3D model of the stroke. Such a representation is drawn in Figure~\ref{fig:3dpose}. \texttt{TTStroke-21} does not offer for the moment any qualitative annotations for strokes. Such quality assessment needs to be built by experts in the field and may be one perspective of such dataset in order to widen its application.

\section{Conclusion}
\label{sec:conclusion}

We have proposed a three-stream network with different kinds of convolutions and input data: raw pixels values and optical flow undergo 3D (2D + time) convolution, while the pose-vectors are submitted to the branch with temporal convolution. Pose information, fused with RGB and optical flow branches, yields much better performances (up to 18\%) in the joint classification and segmentation task. Further analysis may be conducted in this task by developing other evaluation methods not frame-wised.
\par
Improvements can be achieved by developing a better pose estimator which can consider temporal information for avoiding misdetected poses/joints and obtain a better precision of the skeletal joints coordinates. Furthermore, pose information coupled with other technology may be one step forward to gesture analysis in order to assess athletes performance.

\bibliographystyle{ACM-Reference-Format}
\bibliography{biblio}

\end{document}